%% file: main_mlsys_2023.tex
\pdfoutput=1
\documentclass{article}

\usepackage{hyperref}

\usepackage[accepted]{arxivhacked}

\usepackage{amsmath}
\usepackage[inline]{enumitem}
\usepackage{amsmath,graphicx,amsfonts,xcolor,url,booktabs,comment,siunitx}

\usepackage{wrapfig}

\usepackage{pifont}
\newcommand{\cmark}{\text{\ding{51}}}
\newcommand{\xmark}{\text{\ding{55}}}

\usepackage{hyperref}

\newcommand{\norm}[1]{\left\lVert#1\right\rVert}

\usepackage{subfigure}

\usepackage[utf8]{inputenc} 
\usepackage[T1]{fontenc}    
\usepackage{hyperref}       
\usepackage{nicefrac}       
\usepackage{microtype}      

\mlsystitlerunning{FLUTE: A Scalable, Extensible Framework for High-Performance Federated Learning Simulations}

\begin{document}

\twocolumn[
\mlsystitle{FLUTE: A Scalable, Extensible Framework for High-Performance Federated Learning Simulations}



\mlsyssetsymbol{equal}{*}

\begin{mlsysauthorlist}
\mlsysauthor{Mirian Hipolito Garcia}{equal,msr}
\mlsysauthor{Andre Manoel}{equal,msr}
\mlsysauthor{Daniel Madrigal Diaz}{msr}
\mlsysauthor{Fatemehsadat Mireshghallah}{UCSD}
\mlsysauthor{Robert Sim}{msr}
\mlsysauthor{Dimitrios Dimitriadis}{msr}
\end{mlsysauthorlist}

\mlsysaffiliation{msr}{Microsoft Research, Redmond, US}
\mlsysaffiliation{UCSD}{University of California, San Diego, US}

\mlsyscorrespondingauthor{Dimitris Dimitriadis}{didimit@microsoft.com}
\mlsyscorrespondingauthor{Robert Sim}{rsim@microsoft.com}

\mlsyskeywords{federated learning, distributed, privacy preserving ML, simulations, prototyping, platform}

\vskip 0.3in

\begin{abstract}
In this paper we introduce ``Federated Learning Utilities and Tools for Experimentation'' (FLUTE), a high-performance open source platform for federated learning research and offline simulations.  The goal of FLUTE is to enable rapid prototyping and simulation of new federated learning algorithms at scale, including novel optimization, privacy, and communications strategies. We describe the architecture of FLUTE, enabling arbitrary federated modeling schemes to be realized. We compare the platform with other state-of-the-art platforms and  describe available features of FLUTE for experimentation in core areas of active research, such as optimization, privacy, and scalability.  A comparison with other established platforms shows speed-ups of up to $42\times$ and savings in memory footprint of $~3\times$. A sample of the platform capabilities is also presented  for a range of tasks, as well as other functionality, such as linear scaling for the number of participating clients, and a variety of federated optimizers, including FedAdam, DGA, \emph{etcetera}.  
\end{abstract}
]



\printAffiliationsAndNotice{\mlsysEqualContribution} 

\section{Introduction}
\label{sec:intro}

Distributed Training (DT) has drawn much scientific attention with focus on scaling model training processes, either via model or data parallelism. As training datasets grow larger, the need for data parallelism has become a priority. Different approaches have been proposed over the years, as discussed in~\cite{BNH19}, aiming at more efficient training, either in the form of training platforms such as ``Horovod''~\cite{SeBa18, Abadi+16} or algorithmic improvements like ``Blockwise Model-Update Filtering'' (BMUF)~\cite{chen2016scalable}. 
Most often, these techniques and platforms are evaluated on metrics such as data throughput (without compromising model performance), model and  training dataset size, GPU utilization, and convergence rates. There are a few underlying assumptions implied for such DT scenarios, i.e., data and device uniformity and efficient network communication between the working nodes. Besides these specifications, data uniformity is paramount for successful training, ensured by repeated randomization and data shuffling steps.

On the other hand, new constraints in data management are emerging, driven  by the need for privacy compliance for personal data and information~\cite{GDPR}, and widely distributed and segregated data silos. As such, increasingly more data are inaccessible, either behind firewalls or on users' devices without the option of curating for centralized training. The ``\textit{Federated Learning}'' (FL) paradigm has been proposed as a strategy to address these constraints, as in~\cite{McMahan+17}. Federated learning is a decentralized machine learning scheme with a focus on collaborative training and user data privacy. The key idea is to enable training of a global model with the participation of multiple clients coordinated by a central server. Each client trains the model using  local data  and then sends the tuned parameters back to the server, where the global model is updated by aggregating the client local information. 

One of the challenges when using Federated Learning platforms is the need for scaling the learning process to millions of clients, in order to simulate real-world conditions under reasonable computing resources. As such, testing and validating any novel algorithm in realistic circumstances, e.g., using real devices or close-to-real scaled deployments, can be particularly difficult. Simulation platforms play an important role, enabling researchers and developers to develop proof-of-concept implementations (POCs) and validate their performance before building and deploying in production.  While several open-source frameworks have been developed to enable FL solutions, few offer end-to-end simulations, experiment orchestration, and scalability.


This paper introduces a novel platform ``\textit{Federated Learning Utilities and Tools for Experimentation}'' (FLUTE) as a framework for running large-scale, offline FL simulations.  It is designed to be flexible, to enable arbitrary model architectures\footnote{The repository provides some examples 
and users are urged to add their implementations.}, and to allow for prototyping novel approaches in federation, optimization, quantization, privacy, and so on. Finally, it provides an optional integration with AzureML workspaces~\cite{pmlr-v50-azureml15}, enabling scenarios closer to real-world applications, and leveraging platform features to manage and track experiments, parameter sweeps, and model snapshots.

The main contributions of FLUTE are:
\begin{enumerate}[label=(\roman*)]
    \item A platform for high-performance FL simulations at scale (scaling to millions of clients),
    \item Flexibility in the platform to include new FL paradigms, unlocking research, experimentation, and proof-of-concept (PoC) development, 
    \item A generic API for new model types and data formats,
    \item A pre-built list of  features - state-of-the-art federation algorithms, optimizers, differential privacy, bandwidth management, client management/sampling, etc,
    \item Experimental results illustrating the utility of the platform for FL research,
    \item A comprehensive analysis and comparisons with two of the leading FL simulation platforms\footnote{Based on the availability of simulation functionality and the number of downloads from their GitHub repository.}, i.e. FedML~\cite{He+20} and Flower~\cite{Beutel+20}. 
\end{enumerate}

The goal of FLUTE is to facilitate the study of new algorithmic paradigms and optimizations, enabling more effective FL solutions in real-world deployments, by extension the platform has been applied by multiple research groups to advance their ideas, such as:~\cite{dimitriadis20_interspeech},~\cite{https://doi.org/10.48550/arxiv.2207.10308},~\cite{https://doi.org/10.48550/arxiv.2206.00920} and~\cite{https://doi.org/10.48550/arxiv.2210.01834}. FLUTE's unique architecture allows clients to be instantiated on-the-fly once the resource is available and then process each independent client asynchronously, making it more efficient than other platforms. On the other hand, FLUTE does not currently address challenges like data collection, secure aggregation, client-side labeling, or attestation. The code for the platform is open-sourced and available at \textbf{\href{https://github.com/AnonymousQTHM31/FLUTE}{https://github.com/AnonymousQTHM31/FLUTE}}.

\input{prior_work}

\section{FLUTE Platform Design}  
\label{sec:FLUTE}
\input{design}

\section{FLUTE Features}  
\label{app:FLUTE}
FLUTE provides a range of  built-in functionality while state-of-the-art algorithm implementations cover important areas in FL research. In this section we discuss federated optimization, differential privacy, bandwidth efficiency, personalization and computing resource capabilities of FLUTE.


\input{optimizers}

\input{dp}

\input{quantization}

\input{personalization}

\input{AML}

\input{comparison_platform_2}

\input{case_study_2}

\section{Discussion and Conclusions}
\label{sec:conclusions}

In recent years, researchers have made significant efforts to address the challenges Federated Learning (FL), especially when it comes to setting up FL-friendly environments -- privacy guarantees, time-consuming processes, communication costs and beyond. Herein, we  presented FLUTE,  a versatile, open-architecture platform for high-performance federated learning simulation that is available as open source.  
 
 FLUTE's novel architecture provides scaling capabilities, several state-of-the-art federation approaches and related features such as differential privacy and personalization, with a flexible API that enables extensions and the introduction of novel approaches.  FLUTE is model and task-independent, and provides facilities for easy integration of new model architectures based on PyTorch.

The goal of FLUTE is to optimize the available resources, to enable rapid experimentation and prototyping of novel algorithms, facilitating the development of new  FL research efforts in the most expeditious manner.
We encourage the research community to explore new research using FLUTE and invite contributions to the public source repository.

%
%

\bibliographystyle{mlsys2023}

\appendix

%
%

\section{Appendix -- Stale Gradient Analysis}
\label{app:stale_grad}

\input{stale_gradient}
\end{document}

%% file: prior_work.tex
\section{Background and Prior Work}
\label{sec:background}

In general, there are two different approaches concerning the architecture of FL systems: either using a central server~\citep{PaY09}, as the ``coordinator or orchestrator'', or opting for peer-to-peer learning, without the need of a central server~\citep{LJSC20}. 
FLUTE is based on the ``server-client'' architecture, where the server coordinates any number of clients. 
Besides the basic architecture, FLUTE addresses technical challenges such as the required resources, i.e. bandwidth, and computing power, efficiency, optimization and learning pipeline~\citep{SSZ13}, and privacy constraints. 
Such challenges are usually attributed to either the ML side of federated learning, e.g., the distributed nature of the tasks, or the engineering side where the available resources are limited:

\paragraph{Communication overhead:} FL  relies heavily on the communication between server and the clients to complete any training iteration. The fact that some of the clients and the server can be in different networks may cause limited connectivity, high latency, and other failures. Different approaches have been proposed, e.g., gradient quantization and sparsification~\citep{McMahan+17, JGJE21}, different architectures per client~\citep{Cho+22}, use of adapters for federating transformer models, etc. Most of these approaches are already implemented in FLUTE, e.g., quantization results shown in Appendix~\ref{sec:experiments}.

\paragraph{Hardware heterogeneity:} Computing capabilities of each client can vary, i.e., CPU, memory, battery level, storage are not expected to be the same across all nodes. This can affect both the selection and availability of the participating devices and it can bias the learning process. Different approaches have been also proposed to address slower clients, i.e. ``stragglers'', with the most popular allowing for asynchronous updates and client dropouts. 
FLUTE provides a flexible, asynchronous framework to incorporate  workers of different capabilities. Also, there is an intuitive way of modeling faster/slower nodes as part of the training process.

\paragraph{Unbalanced and/or non-IID data:} Local training data are individually generated according to the client usage, e.g., users spending more time on their devices tend to generate more training data than others. Therefore, it is expected that these locally segregated training sets may not be either a representative sample of the global data distribution or uniformly distributed between clients. A simple strategy to overcome communication overheads  was proposed with the ``\textit{Federated Averaging}'' (FedAvg) algorithm~\citep{McMahan+17}. In this approach, the clients perform several training iterations, and then send the updated models back to the server for aggregation based on a weighted average. FedAvg is one of the go-to training strategies for FL, given the simplicity and the consistently good results achieved in multiple experiments. On the other hand, FedAvg is not the best aggregation strategy, especially in the case of non-IID local data distributions. Over time, new approaches have emerged to overcome these limitations, for example, adaptive optimizers~\citep{Reddi+21}, SCAFFOLD~\citep{Karimireddy+20}, and the DGA algorithm~\citep{Dimitriadis+20}, which propose optimization strategies to address the heterogeneity problems on data and devices. 


\paragraph{Threats:} The ever-increasing interest for applying FL in different scenarios has brought interest in malicious attacks and threat models, as described~\citep{LXW22}. FL itself cannot  assure either data privacy, or robustness to diverse attacks proven to be effective in breaking privacy or destroying the learning process. Without any mitigation, both the server and clients can be attacked by malicious users, e.g., attackers can poison the model by sending back to the server fake model parameters \citep{zhang2021} or fake the server and send a malicious model to the clients, stealing local information \citep{enthoven2020}. As such, FL strategies started to incorporate  techniques like Differential Privacy (DP), as detailed in~\citep{wei2020} or Multi-Party Computation (MPC), which only reveals the computation result while maintaining the confidentiality of all the intermediate processes~\citep{byrd2020, Bhowmick+19}. Some defenses against inference and backdoor attacks have been implemented in  FLUTE~\cite{https://doi.org/10.48550/arxiv.2210.01834}.

\paragraph{Simulation and prototyping:} Building federated learning solutions can require significant up-front engineering investment, often with an unclear or uncertain outcome.  Simulation frameworks enable FL researchers and engineers to estimate the potential utility of a particular solution, and investigate novel approaches, before making any significant investments.  Recently, several frameworks have been proposed for FL simulations, including  \textit{TensorFlow Federated}~\citep{Abadi+16}, \textit{FedML}~\citep{He+20} and \textit{Flower}~\citep{Beutel+20}, with each having different focus and different simulation scope of their proof-of-concept scenarios. An extensive comparison of these frameworks, as in Section~\ref{sec:comparison}, shows the competitive advantages of FLUTE.

%% file: design.tex

FLUTE is designed as a scalable framework for rapid prototyping, encouraging researchers to propose novel FL solutions  for real-world applications, in scale, data volume, etc, under the following design constraints/specs:
\begin{itemize}
    \item \textit{Scalability}: Capacity to processs many thousands of clients on any given round. FLUTE allows to run large-scale experiments using any number of clients with reasonable turn-around time, since scale is a critical factor in understanding  practical metrics such as convergence and privacy-utility trade-offs.
    \item \textit{Flexibility}: Allow for any combination of model, dataset and optimizer. FLUTE supports diverse FL configurations, including some of the state-of-the-art algorithms such as DGA~\cite{Dimitriadis+20}, FedAdam~\cite{reddi2020adaptive} and FedAvg~\cite{McMahan+17}, with Pytorch being the framework of choice for implementing the training pipelines.
    \item \textit{Expandable}: Enable end users to incorporate new model architectures and scenarios, and allow them to easily plug in customized/new techniques like differential privacy or gradient quantization, transfer learning and personalization. FLUTE provides an open architecture allowing the incorporation of new algorithms and models in a straightforward fashion. 
\end{itemize}

The current FLUTE implementation provides best-in-class communication speed, outperforming comparable experimental platforms by $<40\times$ speed-ups, see Section~\ref{sec:comparison}.

FLUTE design is based on a central server architecture, as depicted in Figure~\ref{fig:client_server}. The logical workflow is: 
\begin{enumerate*}[label=(\roman*)]
    \item  Send  an initial global model to participating clients,
    \item \label{itm:second} Train instances of the global model with the locally available data on each client,
    \item Send training information, e.g., updated model parameters, logits (if required), and/or gradients/pseudo-gradients back to the server, 
    \item Combine the returned information on the server to produce a new global model,
    \item Optionally, update the global model with an additional server-side  rehearsal step,
    \item \label{itm:sixth} Send  the updated global model to the next sampled subset of clients,
    \item Repeat steps~\ref{itm:second} - \ref{itm:sixth} for the next training iteration.
\end{enumerate*}

%

\input{client_server_image}

\begin{figure*}[ht]
    \centering
    \includegraphics[width=\textwidth]{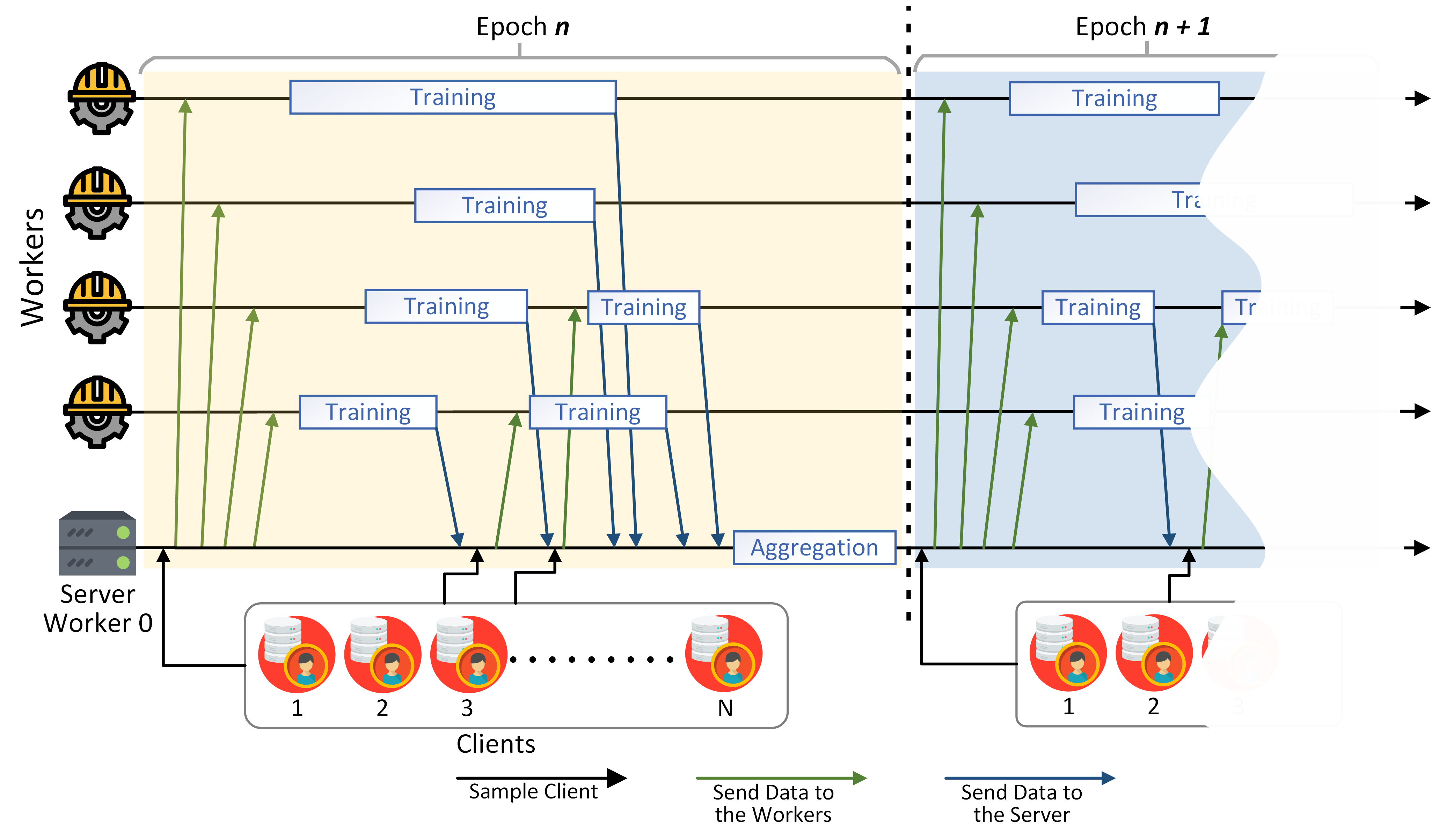}
    \caption{FLUTE Execution Flow: The server samples $\mathcal{N}$ of the clients and sends them to the K workers. Every time one of the workers finishes processing the client data, it returns the gradient and draws the next client until all clients are processed.}
    \label{fig:mpi}
\end{figure*}

A FLUTE job leverages one or more multi-GPU VM running up to a total of $K$ worker processes, each executing tasks assigned by the server (Worker 0). The number of workers is decoupled from the number of clients, $P$, allowing the platform to scale to millions of clients even when $K\ll P$. During training, the server plays the role of both the \emph{orchestrator} and the \emph{aggregator}. First, it distributes a copy of the global model(s), training data and the list of all the client-IDs among the workers. The workers iteratively process the clients' data producing new models, and send the models back to the server. After  $N$ of the clients are processed, the server aggregates the resulting models, typically into a single global model. Algorithm~\ref{alg:flute} describes in detail this process and Figure~\ref{fig:mpi} exemplifies the execution flow.

\begin{algorithm}[ht]
\caption{FLUTE Orchestration: $\mathcal{P}$ is a Client Pool, which contains IDs of each client, $P=|\mathcal{P}|$, $M$ is the federation rounds to be executed,  $K$ is the number of Workers, $\mathcal{N}$ is the subset of clients per iteration and $N=|\mathcal{N}|$ the number of clients per iteration}
\label{alg:flute}
\begin{algorithmic}[l]
    \STATE {\bfseries Server-Side Worker-0:}
    \FOR{ each federated round from $1,\dots,M$ }
        \STATE $\mathcal{N}$ $\leftarrow$ Sample $N$ clients from $\mathcal{P}$
        \REPEAT 
            \STATE Wait for $worker_k$ to finish
            \STATE Save pseudo-gradient response from $worker_k$
            \STATE $c$ $\leftarrow$ Sample client-ID from $\mathcal{C}$
            \STATE Dispatch model and $c$ to $worker_k$
        \UNTIL { all client-IDs $c$ in $\mathcal{N}$ have been processed }
        \STATE Aggregate pseudo-gradients
        \STATE Update model with optimization step
    \ENDFOR
    \STATE 
    \STATE {\bfseries Client-Side Worker-k}, with $\text{k}>0$:
    \STATE Load client and model data
    \STATE Execute local training processes
    \STATE Send pseudo-gradients and statistics about local training to Worker-0
\end{algorithmic}
\end{algorithm}

The distributed nature of FLUTE is based on PyTorch, using the~\texttt{torch.distributed} package as the communication backbone. The interface between server and workers is based on ``\textit{messages}'',  containing model parameters, client-IDs and training instructions. There are four message-types that can be passed from server to worker:
\begin{enumerate*}
  \item \textbf{\texttt{Update}} creates a copy of the model parameters and learning rate on the worker,
  \item \textbf{\texttt{Train}} triggers the execution of a training step on a worker, for a given client. The resulting model (or pseudo-gradient) is passed from worker to server,
  \item \textbf{\texttt{Evaluate}} triggers the execution of an evaluation step in a given client. Resulting metrics are passed from worker to server,
  \item \textbf{\texttt{Terminate}} shuts down the worker thread.
\end{enumerate*}

FLUTE leverages the communication scheme in Figure~\ref{fig:client_server} by loading a local copy of the training data on each worker prior to training,  significantly reducing the traffic communication between server and workers, only sending client indices. In this research simulator, all clients are implemented as isolated object instances, therefore local data on each client stays  within the logical local storage boundaries and it is never  sent to the server or aggregated with other local data sources. Only metrics, model parameters or gradients are communicated between the server and clients -- these quantities can be encrypted~\footnote{Encryption and secure aggregation are not currently implemented in FLUTE - these are security mechanisms which aren't strictly necessary for simulations.} if necessary.

%% file: client_server_image.tex
\begin{figure*}[ht]
    \centering
    \includegraphics[width=0.7\textwidth]{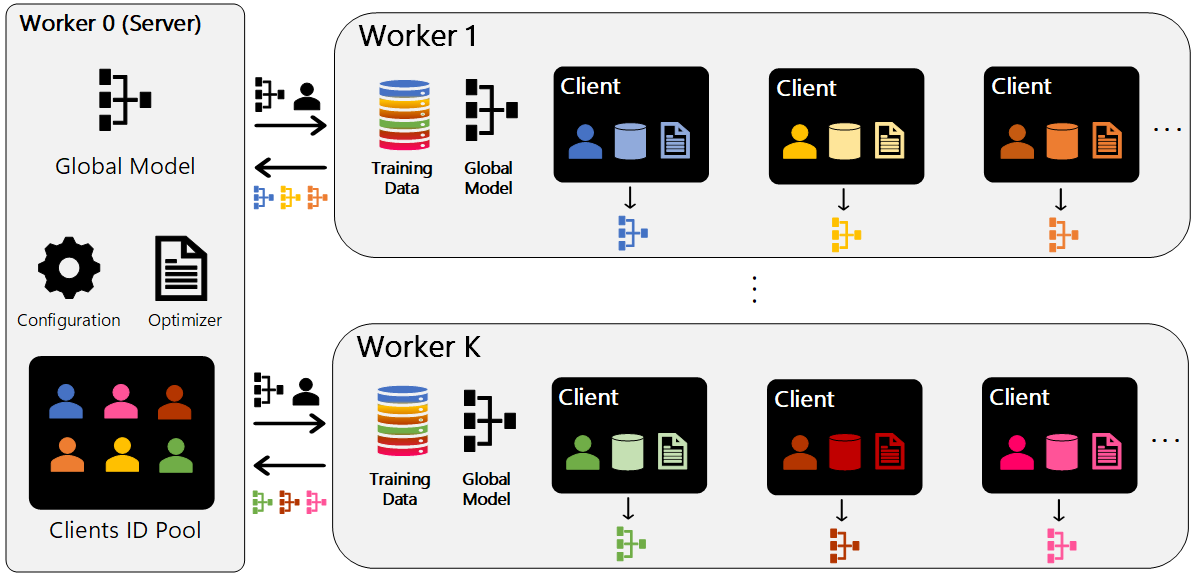}
    \caption{Client-server communication protocol. At each iteration, the server sends a copy of the global model to each worker, samples a number of clients and asynchronously assigns them to a worker as they become available}
    \label{fig:client_server}
\end{figure*}

%% file: optimizers.tex
\subsection{Federated Optimization}
\label{subsec:fed_optimizers}

The ``Federated averaging'' (FedAvg) algorithm,~\cite{KMR15, McMahan+17} is the first and perhaps the most widely used FL training algorithm. The server samples $\mathcal{M}_T\subset \mathcal{N}$  of the available $\mathcal{N}$ devices and sends the model $\textbf{w}^{(s)}_T$ at that current iteration $T$. Each client $j$, $j\in \mathcal{M}_T$ has a version of the model $\textbf{w}^{(j)}_T$, where it is locally updated with the segregated local data. The size of the available data $\mathcal{D}^{(j)}_T$,  per iteration $T$ and client $j$, is expected to differ and as such, $N^{(j)}_T=|\mathcal{D}^{(j)}_T|$ is the size of processed local training samples. After running $\mathcal{E}$ steps of SGD, the updated model $\hat{\textbf{w}}^{(j)}_T$ is sent back to the server. The new model $\textbf{w}_{T+1}$ is given by
\begin{equation}
    \textbf{w}_{T+1} \leftarrow \frac{1}{\sum_{j\in\mathcal{M}_T} {N^{(j)}_T}}\sum_{j\in\mathcal{M}_T}{N^{(j)}_T \hat{\textbf{w}}^{(j)}_T}
\label{eq:fedavg}
\end{equation}
The server-side model in iteration $T+1$ is a weighted average of the locally updated models of the previous iteration. 

Despite some drawbacks like lack of fine-tuning or annealing of a global learning rate,  FedAvg is the baseline aggregation approach in FLUTE, based on its popularity. 

Although the golden standard in FL training, FedAvg presents several drawbacks~\cite{zhao2018federated}. A different family of learning algorithms, ``Adaptive Federated Optimizers'' has been proposed to address  them~\cite{reddi2020adaptive, Dimitriadis+20, LSTS18}, where the clients return pseudo-gradients instead of model parameters.  

\begin{equation}
    \textbf{w}_{T+1} \leftarrow \textbf{w}_{T+1} - f\left(\sum_{j\in\mathcal{M}_T}{\alpha^{(j)} \left(\hat{\textbf{w}}^{(j)}_T-\textbf{w}^{(s)}_{T-1}\right) }\right)
\label{eq:fed_adapt}
\end{equation}
where $f(\cdot)$ is the optimization function, $\hat{\textbf{w}}^{(j)}_T-\textbf{w}^{(s)}_{T-1}$ the difference between the fine-tuned local model and the seed model of the same iteration. Finally, the weights $\alpha$ can be $1$ in the case of FedAdam or take different values, e.g.,  in the case of DGA, as detailed in~\cite{Dimitriadis+21}.

The training process consists of two optimization steps: first, on the client-side using a stateless optimizer for local SGD steps, and then on the server-side with a ``global'' optimizer utilizing the aggregated gradient estimates. The two-level optimization provides both speed-ups in convergence rates due to the second optimizer on the server, and improved control over the training by adjusting the learning rates. In addition, scaling in the number of clients becomes straightforward by adjusting the server-side optimizer. 

The FLUTE system provides support for this group of federated optimizers by adjusting the gradient aggregation weights, and the server-side optimizers, making FedYogi and FedAdam~\cite{reddi2020adaptive}, and DGA~\cite{Dimitriadis+21} rather straightforward to apply. Besides these optimizers, FLUTE provides implementations for FedProx,~\cite{Li+19} and SCAFFOLD,~\cite{Karimireddy+20}. Finally, the FLUTE client scaling capabilities can be enhanced by switching to large-batch optimizers on the server-side, like LAMB~\cite{You+20}, and LARS~\cite{YGG17}, as validated experimentally in Section~\ref{sec:experiments}.

%% file: dp.tex
\subsection{Differential Privacy}
\label{subsec:DP}



In the FL context, differential privacy (DP) \cite{dwork2014algorithmic} is typically enforced by clipping the norm of, and adding noise to, the gradients produced during training \cite{abadi2016deep}. This can be done either by each client (local DP), or by the server (global DP).

In FLUTE, either local or global DP can be used, depending on whether the clients or the server are responsible for doing the clipping and noise-adding. In both cases, that is done directly to the pseudo-gradient, i.e., the difference between current and previous weights after each user's data is processed. The pseudo-gradients are re-scaled so that their norm is at most $C$, ensuring the norm of the difference between any two of them (the sensitivity) is bounded. We typically use Gaussian noise, with variance $\sigma^2 = 2 \log \left(\frac{1.25}{\delta}\right) \frac{C^2}{\epsilon^2}$ picked so that the aggregation is at most $(\epsilon, \delta)$-DP w.r.t. each client. In the case of DGA~\cite{Dimitriadis+20}, the aggregation weights are data-dependent, and also go through the same procedure.

Local clipping and noise addition can be accumulated as if it were per-example (i.e. per-client, or per-pseudo-gradient) clipping and noise addition in global DP-SGD. As such, we track the per-client noise across all clients and apply the Renyi DP accountant globally~\cite{mironov2017renyi}.  Note that tighter bounds have been presented recently in the literature \cite{gopi2021numerical,DBLP:journals/corr/abs-2106-08567} improving these results.

%% file: quantization.tex
\subsection{Bandwidth Efficiency}
\label{subsec:quantization}

Gradients produced during training of the neural networks are known to be quite redundant, and can be compressed without adversely affecting the optimization procedure, e.g., gradient components can be represented by a single bit~\cite{seide20141}, or  some of the components discarded,  making the gradient sparse \cite{wangni2018gradient}. Such compression leads to decreased bandwidth requirements.

In FLUTE, we use an approach similar to that of~\cite{alistarh2017qsgd}. We first get the dynamic range of the gradient components, and then create a histogram of $2^B$ bins between these two values, per layer. Next, we replace each gradient component by the label of the closest bin. That way, only the bin indices, together with the  min./max. values need to be communicated. This quantization procedure is done on the client side, improving the uplink communication.

We can also sparsify the gradients by keeping only the $p\%$  largest (in absolute value) components. If quantization is also active, binning is done before sparsification, but the original value of the component is used to decide whether it is replaced by the bin label or zeroed-out.

As part of FLUTE, novel compression algorithms are proposed, e.g., the use of adapters~\cite{Houlsby+19} when the global models are based on transformers, achieving even further bandwidth compression, or the use of heterogeneous model architectures, as in~\cite{Cho+22}.
In more detail, the transformer models are communicated only once and frozen, while the adapter modules are federated. Despite the significant bandwidth benefits, the computation complexity is increased since all model parameters need to be evaluated. The results are presented in Section \ref{sec:experiments}. 

%% file: personalization.tex
\subsection{Personalization}
\label{subsec:personalization}

The convergence of most  Federated Learning optimization algorithms is  theoretically proven when the client data distributions are iid. However, scenarios where the data distributions are non-iid, are far more challenging, e.g.,~\cite{Li+19}. One of the different approaches for addressing this issue is with convex interpolation between the global $\boldsymbol{\theta}^{(r)}$ and the local models $\boldsymbol{\theta}^{(r,B)}_i$ for the $i^{th}$ client,~\cite{DKMM20}, after $B$ local training epochs. The resulting personalized model $\boldsymbol{\theta}^{(r)}_{int}$ after interpolation is given by    
\begin{equation}
    \boldsymbol{\theta}^{(r)}_{int} = \alpha_i \cdot \boldsymbol{\theta}^{(r, B)}_i+(1-\alpha_i) \cdot \theta^{(r)}
\end{equation}
and the interpolation weights $\alpha_i$ for each client $i$ are estimated as described in~\cite{DKMM20}. FLUTE architecture supports this feature for new models, leveraging clients' local data to obtain models that are better adjusted to the local data distribution. A baseline experiment is included in the repository for computer vision (CV) tasks, as shown in Section~\ref{subsec:personalization_exp}. As expected, a certain drawback of this method is that it requires $P+1$ models for federation with $P$ clients.

%% file: AML.tex
\subsection{Computing Resources}
\label{subsec:aml-integration}

AzureML (AML)~\cite{pmlr-v50-azureml15} is the preferred computing FLUTE environment  for staging, executing, tracking, and summarizing FL experiments. 
FLUTE has native AML integration for job submissions, allowing the users to use the built-in CLI or web interface for job/experiment tracking, and for visualization purposes. While FLUTE needs only the experiment-related configurations,  AML expects the computing environment parameters in a configuration file, such as target, cluster, code, etc. 
Besides AML, FLUTE can also run seamlessly on stand-alone devices, such as laptop and desktop machines (like those used for Section~\ref{sec:comparison}), using the local GPUs if/when available.

%% file: comparison_platform_2.tex
\section{Comparison with related platforms}
\label{sec:comparison}

Multiple FL platforms have been proposed. However, most of them have been designed with a specific purpose limiting their flexibility to experiment with complex large-scale FL scenarios in a reasonable amount of time, and using limited resources. Some production-oriented frameworks also allow researchers to work in a simulation environment using the same platform. Nonetheless, these simulators suffer from lack of flexibility and  limited functionality 
since their architecture is optimized towards productization, especially in complex FL scenarios. Table~\ref{tab:comparison} shows a detailed comparison of the most common FL platforms' features and main focus.

\begin{table*}[t]
\resizebox{\textwidth}{!}{
\centering
    \begin{tabular}{lccc}
        \toprule
         & FLUTE & FedML (Parrot) & Flower (Simulator) \\
        \midrule\midrule
        Focus & Research and Simulation & Research and Production & Research and Production \\
        ML Framework & PyTorch & PyTorch & PyTorch / TensorFlow \\
        Communication Protocols Supported & Gloo,NCCL & SP, MPI & Ray, gRPC \\
        Support Security/Privacy related functions & \cmark & \cmark & \cmark \\
        Support Multiple Federated Optimization Techniques & \cmark & \cmark & \cmark \\
        Flexible and Generic API  & \cmark & \cmark & \cmark \\
        Cloud Integration & \cmark & \cmark & \xmark \\
        Multi-GPU Support & \cmark & \cmark & \xmark \\
        Performance Optimizations & \cmark & \xmark & \xmark \\
        Easily Extensible to Production & \xmark & \cmark & \cmark \\

        \bottomrule
    \end{tabular}
}
\caption{Comparison between FLUTE and Popular Federated Learning Simulation Platforms. This analysis is focused on the simulators provided by these platforms only.}
\label{tab:comparison}
\end{table*}	

FLUTE allows customized training procedures and complex algorithmic implementations at scale, making it a valuable tool to rapidly validate the feasibility of novel FL solutions, while avoiding the need to deal with complications that production environments present.  FLUTE harnesses its architecture to provide a significant advantage in runtime and memory utilization, leveraging benefits of using NCCL~\cite{nvidia} with GPUs, when available.

For this comparison, we selected the \textit{FedML}~\cite{He+20} and \textit{Flower}~\cite{Beutel+20} platforms as the most representative, based on their number of stars on GitHub. Table~\ref{tab:comparison-nccl} shows that FLUTE outperforms FedML by $42\times$ in speed and $3\times$ in memory consumption. The advantage of FLUTE relies on its ability to asynchronously assign new clients to the workers as they become available, and receive their outputs. On the other hand, FedML\footnote{FedML Simulator (Parrot) on its release 0.7.303, commit ID 8f7f261f 
} links the number of workers with the number of MPI processes, which is reflected as the number of parallel clients during training, FLUTE design allows processing multiple clients per worker, decoupling the need for $1:1$ mapping between clients and training processes. In FLUTE, each worker holds a pre-loaded local copy of the training data, avoiding communication overheads during training as the Server only sends indices of the clients to instantiate.

Enabling the comparisons between platforms, the configuration for the experiments are detailed in Table~\ref{tab:nccl-comparison-details}, while all results are shown in Tables~\ref{tab:comparison-nccl} and~\ref{tab:gloo-comparison}. The setups are based on the FedML Benchmarking Recipes\footnote{FedML Benchmarking Results \url{https://doc.fedml.ai/simulation/benchmark/BENCHMARK_MPI.html}} using the same  hyper-parameters, datasets\footnote{FedML Datasets \url{https://github.com/FedML-AI/FedML/tree/master/python/fedml/data}}, and models\footnote{FedML Benchmarking Examples \url{https://github.com/FedML-AI/FedML/tree/master/python/examples/simulation/mpi_fedavg_datasets_and_models_example}}. All FLUTE scripts can be found under \textit{Experiments} in the FLUTE repository. 

\begin{table*}[!ht]
\resizebox{\textwidth}{!}{
\centering
    \begin{tabular}{ccccccccccc}
    \toprule
    Model & Dataset & Algorithm & \# Clients & Clients/round & Batch Size & Client Optim. & lr & Local Epochs & \# Rounds & Test Freq \\
    \midrule\midrule
    Log. Regr. & mnist & FedAvg & 1000 & 10 & 10 & SGD & 0.03 & 1 & 100 & 20\\
    CNN & fedmnist & FedAvg & 3400 & 10 & 20 & SGD & 0.1 & 1 & 800 & 50 \\
    ResNet18 & fedcifar100 & FedAvg & 500 & 10 & 20 & SGD & 0.1 & 1 & 4000 & 50\\
    RNN & fedshakespeare & FedAvg & 715 & 10 & 4 & SGD & 0.8 & 1 & 1200 & 50 \\
    \bottomrule
\end{tabular}
}
\caption{Training configuration for FLUTE/FedML/Flower Benchmarking}
\label{tab:nccl-comparison-details}
\end{table*}


\input{nccl-comparison}

An additional comparison of FLUTE versus Flower 1.0.0~\footnote{Flower Simulator on its release 1.0.0, commit ID 4e7fad9} 
is presented in Table~\ref{tab:gloo-comparison}. The Flower platform seems more efficient when multiple CPUs are employed -- the platform is fairly inefficient by design when multiple GPUs are used  during simulation. To run a fair comparison, we also compare the FLUTE CPU performance (using the Gloo backend) against Flower, evaluating the overall time of the job using the same setup for the \textit{"lr mnist"} task described in Table~\ref{tab:nccl-comparison-details}. FLUTE is  up to $54\times$ faster than Flower on GPUs given that their simulation capabilities are not optimized for multi-GPU jobs. Regardless, FLUTE, running on a Gloo backend, is $9\times$ faster than Flower, running only on CPUs.

\begin{table}[ht]
\centering
\resizebox{0.5\textwidth}{!}{
    \begin{tabular}{ccccc}
    \toprule
    \multicolumn{1}{c}{} &\multicolumn{3}{c}{FLUTE 1.0.0} & {Flower 1.0.0} \\
    \multicolumn{1}{c}{} &\multicolumn{3}{c}{Gloo/NCCL} & {gRPC} \\
    \midrule\midrule
     Accelerator & Acc & Time & Acc & Time\\
    \midrule
    CPU & 80 & \textbf{00:03:20}  & 80 & 00:30:14\\
    GPU 2x & 80 & \textbf{00:01:31}  & 80 & 01:21:44 \\
    GPU 4x & 81 & \textbf{00:01:26}  & 79 & 00:56:45\\
    \bottomrule
\end{tabular}
}
\caption{Performance comparison FLUTE 1.0.0 vs Flower 1.0.0 on 4x NVIDIA RTX A6000, AMD EPYC 7V12 64-Core Processor. Test accuracy is reported from the last communication round.  }
\label{tab:gloo-comparison}
\end{table}

%% file: nccl-comparison.tex
\begin{table*}[!ht]
\resizebox{\textwidth}{!}{
\centering
    \begin{tabular}{llcccccccc}
    \toprule
    \multicolumn{4}{c}{Task} & \multicolumn{3}{c}{FedML (MPI) 0.7.303} & \multicolumn{3}{c}{FLUTE (NCCL) 1.0.0} \\
    \midrule\midrule
     Model & Dataset & Clients & Rounds & Acc & Time & GPU memory & Acc & Time & GPU memory\\
    \midrule
    Log. Regr. & mnist & 1000 & 100 & ~81 & 00:03:09 & ~3060 MB & ~81 & \textbf{00:01:35} & \textbf{~1060 MB} \\
    CNN & fedmnist & 3400 & 800 & ~83 & 05:49:52 & ~5180 MB & ~83 & \textbf{00:08:22} & \textbf{~1770 MB} \\
    ResNet18 & fedcifar100 & 500 & 4000 & ~34 & 15:55:36 & ~5530 MB & ~33 & \textbf{01:42:01} & \textbf{~1900 MB} \\
    RNN & fedshakespeare & 715 & 1200 & ~57 & 06:46:21 & ~3690 MB & ~57 & \textbf{00:21:50} & \textbf{~1270 MB} \\
    \bottomrule
    \end{tabular}
}
\caption{GPU Performance comparison FLUTE 1.0.0 vs FedML 0.7.303 on 4x NVIDIA RTX A6000 using FedML Datasets. Test accuracy  is reported from the last communication round. }
\label{tab:comparison-nccl}
\end{table*}

%% file: case_study_2.tex
\section{Case Studies}
\label{sec:experiments}

This section provides some insights of the FLUTE features. The list of  datasets, tasks and experimental results herein presented is by no means exhaustive. Also, no particular models are detailed since the platform allows training on any architecture currently supported by PyTorch.

\subsection{Sample of Baseline Experiments and Datasets}
\label{subsec:tasks}

We provide  some of the available models/tasks as part of the FLUTE distribution. This list of models and tasks is \textbf{not} exhaustive since the flexibility of the platform allows extensions in models such as Graph Neural Networks (GNNs),~\cite{He+21}, Gradient Boosted Trees (GBTs),~\cite{LWH19}, and others:
\begin{itemize}
\item \textbf{ASR Task: LibriSpeech}:
FLUTE offers a Speech Recognition template task based on the LibriSpeech task~\cite{PCPK15}. The dataset contains about 1,000 hours of speech from 2,500 speakers reading books. Each of the speakers is labeled as a different client. In one of the ASR task examples, a sequence-to-sequence model was used for training, more details can be found in~\cite{Dimitriadis+20b}. 

\item \textbf{Computer Vision Task: MNIST and EMNIST}:
Two different datasets, i.e.,  the MNIST~\cite{mnist2010} and the EMNIST  \cite{cohen2017emnist} dataset are used for Computer Vision tasks. The EMNIST dataset is a set of handwritten characters and digits captured and converted to $28\times28$ pixel images maintaining the image format and data-structure and directly matching the  MNIST dataset. Among the many splits of EMNIST dataset, we use the ``EMNIST Balanced'',  containing \~132k images with 47 balanced classes.

\item \textbf{NLP Tasks: Reddit}:
Various NLP tasks are supported in FLUTE, e.g., 2 use-cases for MLM and next-word prediction using Reddit data~\cite{baumgartner2020pushshift}. The Reddit dataset consists of users' tweets grouped in months as published. For these use-cases, we use 2 months of Reddit data with 2.2M users. 
The seed models used are either from HuggingFace or a baseline LM model, as described below. 

\item \textbf{Sentiment Analysis: sent140, IMDb, YELP}:
Sent140~\cite{sent140}, is a sentiment analysis dataset consisting of tweets, automatically annotated from the emojis found in them. The dataset consists of 255k users, with mean length of 3.5 samples per user. IMDb is based on movie reviews of 1012 users providing 140k reviews with 10 rating classes~\cite{imdb2014}. The YELP dataset is based on restaurant reviews with labels from 1 to 5~\cite{yelp2015}. It contains 2.5k users with 425k reviews.

\item \textbf{Baseline LM Model}: A baseline LM model is used for most of the experiments in Section~\ref{sec:experiments}. A two-layer GRU with 512 hidden units, 10,000 word vocabulary, and embedding dimension 160 is used for fine-tuning during the FL experiments. The seed  model is pretrained on the Google News corpus \cite{Gu+20}.

\end{itemize}


\subsection{Quantization Experiment}
\label{subsec:quant_exp}
\input{quantization_experiments_2}

%

\subsection{Performance for Variable/Different Number of Clients}
\input{scaling.tex}

\subsection{Comparing Optimizers}
\input{opt_expt}

\subsection{Personalization Experiments}
\label{subsec:personalization_exp}

\input{personalization_experiments}

%% file: quantization_experiments_2.tex
The accuracy for a next-word-prediction task on the Reddit dataset and the baseline LM model (as described in Section~\ref{subsec:tasks}) for various levels of quantization $B$ is shown in Table~\ref{tab:quant}.  As expected, using less bits leads to decreased performance in terms of accuracy.
\begin{table}[!ht]
\centering
\resizebox{0.5\textwidth}{!}{
\begin{tabular}{lccc}
\toprule
 & Quant. (bits) & Acc @1 (\%) & Rel. Imprv. (\%) \\
\midrule\midrule
Seed Model          & N/A           &  9.83 & (56.62) \\
Server-Side Training & N/A           & 22.30 & (1.59)        \\
\midrule
FL Train.           & 32            & 22.70 & 0 \\
                    & 10            & 22.40 & (1.32) \\
                    &  8            & 22.20 & (2.25) \\
                    &  4            & 21.30 & (5.87) \\
                    &  3            & 18.80 & (17.21) \\
                    &  2            & 17.80 & (21.58) \\
\bottomrule
\end{tabular}
}
\caption{Next-word prediction: Top-1 accuracy after gradient quantization. The number of bits per gradient coefficient varies $2-32$.}
\label{tab:quant}
\end{table}

We also experiment with the sparsity level, while keeping the quantization set at 8 bits, cf. Table~\ref{tab:sparse}. Herein, we observe gains in bandwidth of up to $16\times$ with no significant change in performance. Error compensation techniques, e.g., \cite{strom2015scalable} could be used to increase the performance at higher sparsity levels. The differences for the case of 8-bit quantization level in Tables~\ref{tab:quant} and~\ref{tab:sparse} are due to noise during the training process. 
\begin{table}[!ht]
\centering
\begin{tabular}{ccc}
\toprule
 \% Sparsity & Gain in Bandwidth & Acc @1 (\%) \\
\midrule\midrule
0.0 & 4x & 22.60 \\
75.0 & 16x & 21.70 \\
95.0 & 80x & 19.00 \\
99.0 & 400x & 17.70 \\
\bottomrule
\end{tabular}
\caption{Performance obtained by varying sparsity level on gradients while keeping quantization fixed at 8 bits -- gains in bandwidth are relative to standard 32 bits gradient. The performance reported is the best over 5000 iterations, with 1000 clients/iteration.}
\label{tab:sparse}
\end{table}

Next, we showcase the efficiency of federating adapters rather than the full transformer models on the sent140 task. The dataset contains about $255k$ unique user-IDs with a total number of $~1.19M$ sentences and very skewed distribution of sentences per user, i.e. average number of sentences per user is $3.56$. The base model is a \textit{BeRT-base-uncased} transformer with $109M$ parameters and the adapters have only $900k$ parameters. We use two version of the adapters, either pretrained (provided by HuggingFace) or randomly initialized.  The 3 different federation strategies are:~(i) federating the entire model,~(ii) federating the last layer of $7.68M$ parameters, and~(iii) federating the adapter components with total size of $800k$ parameters. The non-iid scenario is based on the actual user data distribution, i.e. every user-ID is a separate client. In the case of the iid data distribution, we are creating clients by concatenating together user data.
\begin{table}[!ht]
\center
\label{tab:adaptors}
\resizebox{0.5\textwidth}{!}{
\begin{tabular}{lccc}
\toprule
Setup & Federation Strategy & Gain in Bandwidth & Acc (\%)  \\
\midrule\midrule
Centralized          & Full-model               & NA     &  86.9 \\
Training             & Pretrained Adapters     & NA     &  86.6 \\
\midrule
FL Train.           & Full-model                & 0.0    & 83.5 \\
(non-iid)           & Last-layer                &  $14.2\times$ & 83.1  \\
                    & Adapters (rand. init.)    &  $121\times$  & 83.2  \\
                    & Pretrained Adapters      &  $121\times$  & 83.4  \\
\midrule
FL Train.           &  Full-model               & 0.0    & 86.1 \\
(iid)               &  Pretrained Adapters     & $121\times$   & 86.1 \\
\bottomrule
\end{tabular}
}
\caption{Fine-tuning Strategy: Federating~(a) entire model,~(b) last layer, and~(c) adapter modules. All results are sentiment classification accuracy.}
\end{table}
The observed performance degradation is due to the skewed data distribution per client. Further, some of the available clients don't have enough data for a full batch and we employ  zero-padding. Federating just the adapters can achieve as good or better performance as federating the full models, as shown in Table~\ref{tab:adaptors}. In such case, the gains in bandwidth utilization is about $121\times$ without any loss in performance.

%% file: scaling.tex
The number of clients processed at each round is a variable we can control on FLUTE. Here, we show in Table~\ref{tab:scaling_expt} a simulation with 1 server + 3 workers attached to RTX A6000 GPUs and 2.45GHz AMD EPYC cores for varying number of clients per iteration. Since clients are processed sequentially by each worker, runtime scales linearly. FLUTE also provides options for further speed-ups by processing clients in multiple threads and pre-encoding the data.

\begin{table}[!ht]
    \centering
    \begin{tabular}{cc}
    \toprule
    Number of Clients & Runtime (sec.) \\ \hline\hline
    1,000 & 22.1 $\pm$ 0.6 \\
    5,000 & 111.3 $\pm$ 2.4 \\
    10,000 & 219.0 $\pm$ 2.3 \\
    50,000 & 1103.7 $\pm$ 11.3 \\
    \bottomrule
    \end{tabular}
    \caption{How long it takes for 3 workers to process different number of clients, on an NLG experiment using a GRU model and the Reddit dataset. Averages are computed over 20  iterations.}
    \label{tab:scaling_expt}
\end{table}

Table~\ref{tab:scaling_expt} shows that FLUTE scales gracefully the number of clients per iteration, without any upper bound to that number. We can also look at the predictive performance attained for different numbers of clients, and study how it changes as a function of the optimizer used.

\begin{table}[!ht]
    \resizebox{0.5\textwidth}{!}{
    \centering
    \begin{tabular}{ccc}
    \toprule
    Num. of Clients & Optimizer     & Acc @ 1 (\%) \\
                    & local-server  &  \\ \hline\hline
    No Fine-tuning  & Seed Model & 9.80 \\ \hline
    1k clients/iter & SGD-Adam (Baseline) & 22.70 \\
                    & SGD-RL-based DGA  & 22.80 \\ \hline
    10k clients/iter& SGD-Adam (Baseline) & 20.80 \\
                    & SGD-LARS & 17.00 \\
                    & Adam-LARS & 21.40 \\
                    & SGD-LAMB & 23.00 \\ \hline
    Variable number & SGD-Adam & 22.30 \\
    $[5k-10k]$ clients/iter &  &  \\
    \bottomrule
    \end{tabular}
    }
    \caption{Next-word Prediction task: Top-1 accuracy achieved varying number of clients and optimizers.}
    \label{tab:scaling_exp_opt}
\end{table}

In Table \ref{tab:scaling_exp_opt} , we compare 4 different scenarios for optimizers, increasing the number of clients, showing that the accuracy remains stable for most of them. However, the Adam optimizer decreases its accuracy as the number of clients increase, compared to SGD-LAMB that reaches a better performance with a larger number of clients.

%% file: opt_expt.tex
This experiment of next-word prediction, using the Reddit dataset and baseline LM model described in Section~\ref{subsec:tasks}, explores model training performance for a variety of state-of-the-art optimizer choices.  We trained a recurrent language model, fixing the number of clients per round to 1,000, and varying the choice of optimizer in the central aggregator.  Specifically, we applied standard SGD~\cite{Rosenblatt58}, ADAM~\cite{KiBa17}, LAMB~\cite{You+20}, and LARS~\cite{YGG17}. Table~\ref{tab:opt_expt} illustrates the performance of each optimizer, including maximum validation accuracy, and convergence rate: the number of rounds to reach 95\% of the max. accuracy. Note there is no hyper-parameter tuning of the optimizers for this experiment.

\begin{table}[htb]
    \centering
    \begin{tabular}{lcr}
    \toprule
        Optimizer & Acc @1 (\%) & Convergence Round  \\ \hline\hline
        LAMB & 23.10 & 115 \\
        ADAM & 22.70 & 641 \\
        SGD &  20.60 &  2172\\
        LARS & 17.40 &  414\\
        \bottomrule
    \end{tabular}
    \caption{Next-word prediction task: Top-1 Accuracy and training rounds to 95\% convergence for various central optimizer choices.}
    \label{tab:opt_expt}
\end{table}

%% file: personalization_experiments.tex
The CIFAR-10 task is used for the personalization experiments, splitting the data across 100 clients, and sampling 10 clients per iteration. The client data are split according to the process described in~\cite{He+20}, with $\alpha\in[0.2,\ 1.0]$ for the Dirichlet label distribution (client distributions are more iid when the $\alpha$ values are larger). In addition to the label distributions, we investigate different feature distributions by applying locally different image transformations (per each client). For this experiment, we fix the test samples to match the local training data/label distributions, i.e. we split the test set  to follow similar local label distributions as the training samples. The image transformations are unique per client for both the training and test samples, when applicable. Herein, we investigate 3 different training strategies, i.e. a global model trained with DGA, local models trained with SGD and the convex interpolation of these two, as described in Section~\ref{subsec:personalization}. 
The relative performance improvement shown in Table~\ref{tab:personalization} is between the global and the interpolated models.
\begin{table}[ht]
\centering
\resizebox{0.5\textwidth}{!}{
\label{tab:personalization}
\begin{tabular}{lcccc}
\toprule
                         & Global & Local & Interp. & Rel. Imprv.\\
\midrule\midrule
iid  ($\alpha=1.0$)     & 74.12 & 46.10 & 77.72 & \textbf{13.91} \\
non-iid  ($\alpha=0.5$) & 72.33 & 54.90 & 79.56 & \textbf{26.13} \\
non-iid  ($\alpha=0.2$) & 69.50 & 70.70 & 85.43 & \textbf{52.23} \\
\midrule
\midrule
iid  ($\alpha=1.0$)     & 51.34 & 46.48 & 62.78 & \textbf{23.51} \\
\ \ + Feat. Transf. & & & & \\
non-iid  ($\alpha=0.5$) & 49.40 & 54.57 & 67.60 & \textbf{35.97} \\
\ \ + Feat. Transf. & & & & \\
non-iid  ($\alpha=0.2$) & 47.55 & 70.90 & 77.45 & \textbf{57.01} \\
\ \ + Feat. Transf. & & & & \\
\bottomrule
\end{tabular}
}
\caption{Personalization on CIFAR-10: Two sources of non-iidness,~(i) Label distribution based on $\alpha\in[0.2,\ 1.0]$ and~(ii) Different image transformations per client. All reported results are in image classification accuracy (\%).}
\end{table}
Convex interpolation, as described in Section~\ref{subsec:personalization}, always benefits overall performance but the gains are more significant in the case of extreme non-iidness. 

Increasing the local non-iidness helps the \textit{local} model performance (and the interpolated combination between local and global models). The local datasets have more examples of the particular labels since the total number of local training samples remains constant. As such, the local models can generalize better, improving the overall performance.

The feature-based non-iidness doesn't affect the local models, since these models are trained on matched transformations, their impact on the model quality is minimal. On the contrary, image transformations have great impact on the global models due to the increased  data mismatch.

%% file: stale_gradient.tex
The FLUTE platform offers flexibility when sampling the participating clients from the pool of candidates. A range in the number of clients can be given, fluctuating between the two ends. This can be seen as ``\textit{client dropouts}'', where a number of clients can be randomly discarded. Since the default optimization pipeline is based on DGA, as in Section~\ref{subsec:fed_optimizers}, the learning rate can be  adjusted accordingly.

\begin{figure}[!t]
    \centering
    \includegraphics[width=0.5\textwidth]{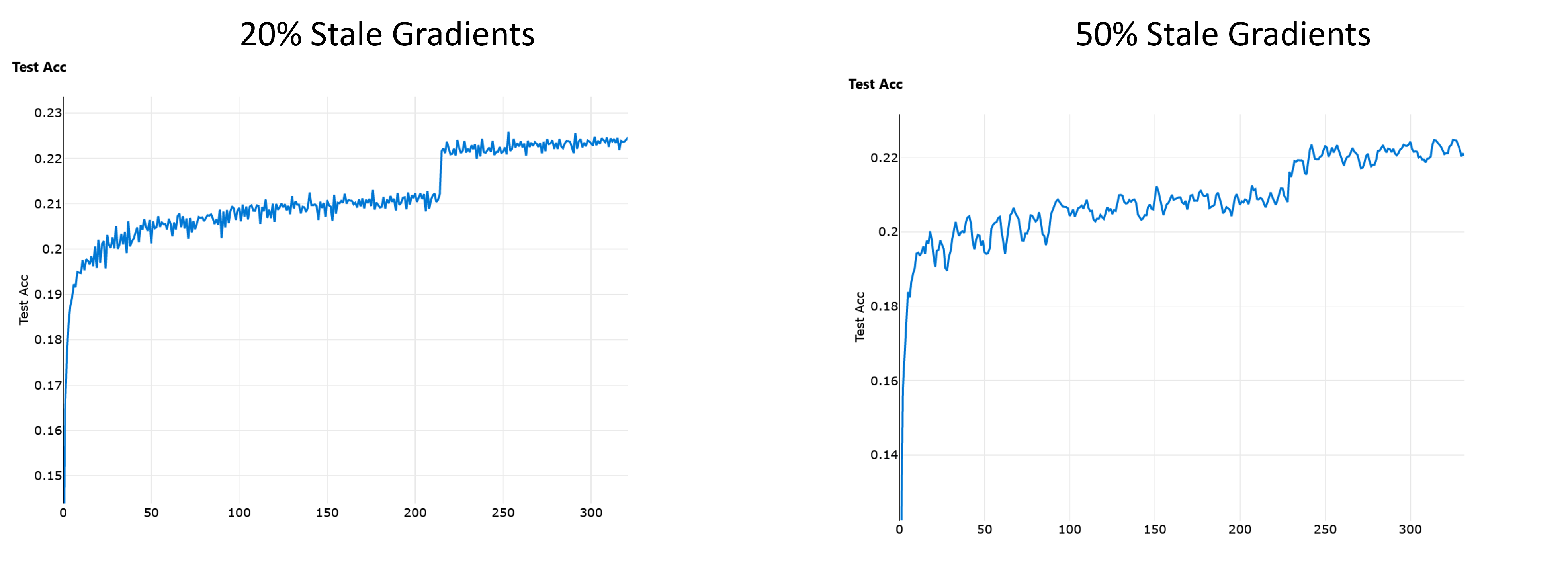}
    \caption{Next-word Prediction task: Top-1 Accuracy for Reddit dataset with Staleness or 1 iteration.}
    \label{fig:stale}
\end{figure}

Similar to the dropout functionality, FLUTE offers an option of delaying the contributions of random clients rather than discarding the corresponding gradients. The system can introduce a 1-step ``staleness'' to the system by randomly delaying a subset of the clients by 1 iteration. The convergence analysis for the stale gradients scenario is held in  Appendix~\ref{app:stale_grad}. As shown, the error introduced due to staleness is upper bounded. As such, there is theoretical guarantee that the model will finally converge. The theoretical conclusions are experimentally verified using FLUTE, shown in Figure~\ref{fig:stale}.

A complementary approach to deal with the issue of straggling is to use asynchronous SGD. In asynchronous SGD, any learner can evaluate the gradient and update the central PS without waiting for the other learners. Asynchronous variants of existing SGD algorithms have also been proposed and implemented in systems, e.g.,~\cite{AgDu11, Dutta+18}. In general, analyzing the convergence of asynchronous SGD with the number of iterations is difficult in itself because of the randomness of gradient staleness.

 Gradient descent is a way to iteratively minimize this objective function by updating the parameter $w$ based on the gradient of the model $\mathbf{\theta}^{(s)}_{\tau}$ at every iteration $\tau$, as given by
\begin{equation}
\mathbf{\theta}^{(j)}_{t+1}=\mathbf{\theta}^{(j)}_{t}-\eta^{(j)}\nabla_\theta\mathcal{L}_\theta(\mathbf{x}^{(j)}_i)
 \label{eq:sgd}
\end{equation}
for the $j$ client, over the local data mini-batches $\mathbf{x}^{(j)}_i$.
As described in Section~\ref{subsec:fed_optimizers} the clients are estimating a pseudo-gradient $\mathbf{\tilde{g}}^{(j)}_{T_j+\tau}$ at the end of their training cycle,
\begin{equation}
\mathbf{\tilde{g}}^{(j)}_{T_j+\tau}=\mathbf{\theta}^{(j)}_{T_j}-\mathbf{\theta}^{(s)}_{\tau}
 \label{eq:pseudo_grad_time}
\end{equation}
where $T_j$ is the time took for the client $j$ to estimate the final local model, and $\theta_\tau$ is the global/initial model communicated to the client at time $\tau$. As in~\cite{Dimitriadis+20}, these pseudo-gradients are weighted and aggregated
\begin{equation}
\mathbf{\theta}^{(s)}_{\tau+1} = \mathbf{\theta}^{(s)}_\tau- \eta^{(s)} \sum_{j\in N}{\mathbb I _{j,\tau} \alpha_j \mathbf{\tilde{g}}^{(n)}_{T_j+\tau}}
 \label{eq:dga_time}
\end{equation}
where $N$ the number of clients per iteration $\tau$, and the samples of 
\[
    \mathbb I _{j,\tau}= 
            \left\{
        	\begin{array}{ll}
        		1 & \mbox{ if \ } T_j+\tau \in W_{[\tau, \tau+1)} \\
        		0 & \mbox{ else}
        	\end{array}
        \right.
\] 
 and $\widetilde{\mathbb I} _{j,\tau}=1-\mathbb I _{j,\tau}$

There are different degrees of staleness and for this work, the stale gradients are considered to fall at most one iteration behind, i.e. some of the gradients $\tilde{g}^{(j)}_{T_j+\tau-1}$ are part of the aggregation step in Eq.~\ref{eq:dga_time} for the window $W_{[\tau, \tau+1)}$. In other words, Eq.~\ref{eq:dga_time} now becomes
\begin{equation}
\resizebox{0.45\textwidth}{!}{
$\mathbf{\theta}^{(s)}_{\tau+1} = \mathbf{\theta}^{(s)}_\tau- \eta^{(s)} 
\left[ \sum_{j\in J}{\alpha_j (\mathbf{\theta}^{(j)}_{T_j}-\mathbf{\theta}^{(s)}_{\tau})}+ \sum_{i\in I}{\alpha_i (\mathbf{\theta}^{(i)}_{T_i}-\mathbf{\theta}^{(s)}_{\tau-1})}\right]$
}
\label{eq:dga_stale}
\end{equation}
where $J,\ I$ is the index of nodes without/with stale gradients and assuming that $J\cup I= N$, i.e, the union of clients with current and stale gradients cover the client space per iteration. Assuming that the final models $\theta_{T_j}$ per client, would reach a similar point regardless of the starting model $\theta^{(s)}_\tau$ (a realistic assumption in convex models).
\begin{equation}
\resizebox{0.45\textwidth}{!}{
$    \mathbf{\theta}^{(s)}_{\tau+1} \approx \mathbf{\theta}^{(s)}_\tau- \eta^{(s)} 
    \left[ \sum_{n\in N}{\alpha_n (\mathbf{\theta}^{(n)}_{T_n}-\mathbf{\theta}^{(s)}_{\tau})}+ \sum_{i\in I}{\alpha_i (\mathbf{\theta}^{(s)}_\tau-\mathbf{\theta}^{(s)}_{\tau-1})}\right]$
}
\label{eq:dga_stale_2}
\end{equation}

Based on Eqs.~\ref{eq:dga_time}, \ref{eq:dga_stale_2}, the stale gradients of the $I$ nodes introduces an error term $E_\tau$ which depends only on the weights $\alpha_i$ and the difference with the previous model, i.e, the aggregated gradients of the previous time-step,
\[
\begin{split}
    E_\tau   = \eta^{(s)}\left(\mathbf{\theta}^{(s)}_\tau-\mathbf{\theta}^{(s)}_{\tau-1}\right) & \sum_{i\in I}{\alpha_i } \\
      & =\eta^{(s)}\left(\mathbf{\theta}^{(s)}_\tau-\mathbf{\theta}^{(s)}_{\tau-1}\right)\sum_{i\in I}{\widetilde{\mathbb I} _{i,\tau}\alpha_i }
\end{split}
\]


The expectation of the $L2$-norm of the error is, 
\begin{equation}
    \begin{split}
        \mathbb{E}\left[\norm{E_\tau}_2\right] &= \\
        &= \mathbb{E}\left[\norm{\eta^{(s)}\left(\mathbf{\theta}^{(s)}_\tau-\mathbf{\theta}^{(s)}_{\tau-1}\right)\sum_{i\in I}{\widetilde{\mathbb I} _{i,\tau}\alpha_i }}_2\right] \\
        &\leq  \eta^{(s)} \mathbb{E}\left[\norm{\left(\mathbf{\theta}^{(s)}_\tau-\mathbf{\theta}^{(s)}_{\tau-1}\right)}_2\right]
    \end{split}
\label{eq:upper_bound}
\end{equation}
since $\sum{\widetilde{\mathbb I} _{i,\tau}\alpha_i }\leq 1$. According to Eq.~\ref{eq:upper_bound}, the upper-bound of the error term due to the stale gradients is the norm of the model differences between updates weighted by the learning rate $\eta^{(s)}$. In other words, the expectation of the norm of the error due to stale gradients is bound by the model updates (in fixed points in time). 
If we call $\Delta_\tau$ the norm of the difference between sequential in time models, 
\begin{equation}
    \Delta_\tau = \norm{\mathbf{\theta}^{(s)}_\tau-\mathbf{\theta}^{(s)}_{\tau-1}}_2
\label{eq:diff_models}
\end{equation}
becomes smaller since the models converge to an optimal point. As such, $\lim_{\tau \to\infty} \Delta_\tau =0$ and from Eq.~\ref{eq:upper_bound}, the error due to stale gradients becomes $\lim_{\tau \to\infty} E_\tau =0$.

The conclusion from  Eqs.~\ref{eq:upper_bound} and~\ref{eq:diff_models} is in accordance with the analysis in~\cite{LHLL15}, where it is shown that the convergence rate does not depend on the staleness ratio given sufficient number of iterations. It is proved that the benefits of not waiting for the strangler nodes (thus producing stale gradients) in terms of time needed to converge counter-balance the errors introduced early in the training process. Also, based on the analysis in~\cite{Dutta+18}, adjusting the learning rate schedule per iteration $\tau$ based on the staleness $\Delta_\tau$ can further expedite convergence,
\begin{equation}
    \eta^{(s)}_\tau=\min \left\{ \frac{C}{\Delta_\tau},\ \eta_{\max} \right\},
    \label{eq:lr_schedule_staleness}
\end{equation}
where $C$ is a predefined constant related to the error floor.


%

We experimentally verify the theoretical analysis in Appendix~\ref{app:stale_grad} with two different experiments, depending on the percentage of stale clients, 20\% and 50\% of the 1000 clients are stale -- staleness in this experiment equals to 1 cycle. This experiment is based on the next-word prediction task using the Reddit dataset, together with the baseline LM model described in Section~\ref{subsec:tasks}. As suggested, the model still converges to an optimal point in terms of accuracy. However, it takes longer for the case of 50\% to reach a good point in performance. 

%% file: main_mlsys_2023.bbl
\begin{thebibliography}{63}
\providecommand{\natexlab}[1]{#1}
\providecommand{\url}[1]{\texttt{#1}}
\expandafter\ifx\csname urlstyle\endcsname\relax
  \providecommand{\doi}[1]{doi: #1}\else
  \providecommand{\doi}{doi: \begingroup \urlstyle{rm}\Url}\fi

\bibitem[Abadi et~al.(2016{\natexlab{a}})Abadi, Barham, Chen, Chen, Davis,
  Dean, Devin, Ghemawat, Irving, Isard, Kudlur, Levenberg, Monga, Moore,
  Murray, Steiner, Tucker, Vasudevan, Warden, Wicke, Yu, and Zheng]{Abadi+16}
Abadi, M., Barham, P., Chen, J., Chen, Z., Davis, A., Dean, J., Devin, M.,
  Ghemawat, S., Irving, G., Isard, M., Kudlur, M., Levenberg, J., Monga, R.,
  Moore, S., Murray, D., Steiner, B., Tucker, P., Vasudevan, V., Warden, P.,
  Wicke, M., Yu, Y., and Zheng, X.
\newblock {Tensorflow: A System for Large-Scale Machine Learning}.
\newblock \emph{arXiv preprint arXiv:1605.08695}, 2016{\natexlab{a}}.

\bibitem[Abadi et~al.(2016{\natexlab{b}})Abadi, Chu, Goodfellow, McMahan,
  Mironov, Talwar, and Zhang]{abadi2016deep}
Abadi, M., Chu, A., Goodfellow, I., McMahan, H.~B., Mironov, I., Talwar, K.,
  and Zhang, L.
\newblock Deep learning with differential privacy.
\newblock In \emph{Proceedings of the 2016 ACM SIGSAC Conf. on computer and
  communications security}, pp.\  308--318, 2016{\natexlab{b}}.

\bibitem[Agarwal \& Duchi(2011)Agarwal and Duchi]{AgDu11}
Agarwal, A. and Duchi, J.~C.
\newblock Distributed delayed stochastic optimization.
\newblock \emph{Advances in Neural Information Processing Systems}, pp.\
  873–--881, 2011.

\bibitem[Alistarh et~al.(2017)Alistarh, Grubic, Li, Tomioka, and
  Vojnovic]{alistarh2017qsgd}
Alistarh, D., Grubic, D., Li, J., Tomioka, R., and Vojnovic, M.
\newblock Qsgd: Communication-efficient sgd via gradient quantization and
  encoding.
\newblock \emph{Advances in Neural Information Processing Systems},
  30:\penalty0 1709--1720, 2017.

\bibitem[{AzureML Team}(2016)]{pmlr-v50-azureml15}
{AzureML Team}.
\newblock {AzureML: Anatomy of a Machine Learning service}.
\newblock In \emph{Proc. of The 2nd Intern. Conf. on Predictive APIs and Apps},
  Proc. of Machine Learning Research, pp.\  1--13. PMLR, Aug. 2016.

\bibitem[Baumgartner et~al.(2020)Baumgartner, Zannettou, Keegan, Squire, and
  Blackburn]{baumgartner2020pushshift}
Baumgartner, J., Zannettou, S., Keegan, B., Squire, M., and Blackburn, J.
\newblock The pushshift reddit dataset.
\newblock In \emph{Proc. of the Intern. AAAI Conf. on web and social media},
  volume~14, pp.\  830--839, 2020.

\bibitem[Ben-Nun \& Hoefler(2019)Ben-Nun and Hoefler]{BNH19}
Ben-Nun, T. and Hoefler, T.
\newblock {Demystifying Parallel and Distributed Deep Learning: An In-depth
  Concurrency Analysis}.
\newblock \emph{ACM Computing Surveys}, 4\penalty0 (65), 2019.

\bibitem[Beutel et~al.(2020)Beutel, Topal, Mathur, Qiu, Fernandez-Marques, Gao,
  Sani, Li, Parcollet, de~Gusmão, and Lane]{Beutel+20}
Beutel, D.~J., Topal, T., Mathur, A., Qiu, X., Fernandez-Marques, J., Gao, Y.,
  Sani, L., Li, K.~H., Parcollet, T., de~Gusmão, P. P.~B., and Lane, N.~D.
\newblock Flower: A friendly federated learning research framework.
\newblock \emph{arXiv preprint arXiv:2007.14390}, 2020.

\bibitem[Bhowmick et~al.(2019)Bhowmick, Duchi, Freudiger, Kapoor, and
  Rogers]{Bhowmick+19}
Bhowmick, A., Duchi, J., Freudiger, J., Kapoor, G., and Rogers, R.
\newblock Protection against reconstruction and its applications in private
  federated learning.
\newblock \emph{arXiv preprint arXiv:1812.00984}, 2019.

\bibitem[Byrd \& Polychroniadou(2020)Byrd and Polychroniadou]{byrd2020}
Byrd, D. and Polychroniadou, A.
\newblock Differentially private secure multi-party computation for federated
  learning in financial applications.
\newblock \emph{arXiv preprint arxiv:2010.05867}, 2020.

\bibitem[Chen \& Huo(2016)Chen and Huo]{chen2016scalable}
Chen, K. and Huo, Q.
\newblock Scalable training of deep learning machines by incremental block
  training with intra-block parallel optimization and blockwise model-update
  filtering.
\newblock In \emph{Proc. ICASSP}, March 2016.

\bibitem[Cho et~al.(2022)Cho, Manoel, Joshi, Sim, and Dimitriadis]{Cho+22}
Cho, Y.~J., Manoel, A., Joshi, G., Sim, R., and Dimitriadis, D.
\newblock Heterogeneous ensemble knowledge transfer for training large models
  in federated learning.
\newblock In \emph{Proc. of the 31st Intern. Joint Conf. on Artificial
  Intelligence, {IJCAI-22}}, pp.\  2881--2887. Intern. Joint Conf. on
  Artificial Intelligence Org., Juky 2022.

\bibitem[Cohen et~al.(2017)Cohen, Afshar, Tapson, and
  Van~Schaik]{cohen2017emnist}
Cohen, G., Afshar, S., Tapson, J., and Van~Schaik, A.
\newblock Emnist: Extending mnist to handwritten letters.
\newblock In \emph{Intern. Joint Conf. on Neural Networks (IJCNN'17)}, pp.\
  2921--2926. IEEE, 2017.

\bibitem[Deng et~al.(2020)Deng, Kamani, and Mahdavi]{DKMM20}
Deng, Y., Kamani, M.~M., and Mahdavi, M.
\newblock Adaptive personalized federated learning.
\newblock \emph{arxiv preprint arXiv:2003.13461}, 2020.

\bibitem[Diao et~al.(2014)Diao, Qiu, Wu, Smola, Jiang, and Wang]{imdb2014}
Diao, Q., Qiu, M., Wu, C.-Y., Smola, A.~J., Jiang, J., and Wang, C.
\newblock Jointly modeling aspects, ratings and sentiments for movie
  recommendation (jmars).
\newblock In \emph{Proc. of the 20th ACM SIGKDD Intern. Conf. on Knowledge
  Discovery and Data Mining}, 2014.

\bibitem[Dimitriadis et~al.(2020{\natexlab{a}})Dimitriadis, Kumatani, Gmyr,
  Gaur, and Eskimez]{Dimitriadis+20}
Dimitriadis, D., Kumatani, K., Gmyr, R., Gaur, Y., and Eskimez, E.~S.
\newblock {Federated Transfer Learning with Dynamic Gradient Aggregation}.
\newblock \emph{arXiv preprint arXiv:2008.02452}, 2020{\natexlab{a}}.

\bibitem[Dimitriadis et~al.(2020{\natexlab{b}})Dimitriadis, Kumatani, Gmyr,
  Gaur, and Eskimez]{Dimitriadis+20b}
Dimitriadis, D., Kumatani, K., Gmyr, R., Gaur, Y., and Eskimez, S.~E.
\newblock A federated approach in training acoustic models.
\newblock In \emph{In Proceedings of Interspeech'20}, 2020{\natexlab{b}}.

\bibitem[Dimitriadis et~al.(2020{\natexlab{c}})Dimitriadis, Kumatani, Gmyr,
  Gaur, and Eskimez]{dimitriadis20_interspeech}
Dimitriadis, D., Kumatani, K., Gmyr, R., Gaur, Y., and Eskimez, S.~E.
\newblock {A Federated Approach in Training Acoustic Models}.
\newblock In \emph{Proc. Interspeech 2020}, pp.\  981--985, 2020{\natexlab{c}}.
\newblock \doi{10.21437/Interspeech.2020-1791}.

\bibitem[Dimitriadis et~al.(2021)Dimitriadis, Kumatani, Gmyr, Gaur, and
  Eskimez]{Dimitriadis+21}
Dimitriadis, D., Kumatani, K., Gmyr, R., Gaur, Y., and Eskimez, S.~E.
\newblock {Dynamic Gradient Aggregation for Federated Domain Adaptation}.
\newblock \emph{arXiv preprint arXiv:2106.07578}, 2021.

\bibitem[Dutta et~al.(2018)Dutta, Joshi, Ghosh, P., and P.]{Dutta+18}
Dutta, S., Joshi, G., Ghosh, S., P., D., and P., N.
\newblock Slow and stale gradients can win the race: Error-runtime trade-offs
  in distributed sgd.
\newblock In \emph{Proc. of the 21st Intl. Conf. on Artificial Intelligence and
  Statistics}, pp.\  803--812, 2018.

\bibitem[Dwork et~al.(2014)Dwork, Roth, et~al.]{dwork2014algorithmic}
Dwork, C., Roth, A., et~al.
\newblock The algorithmic foundations of differential privacy.
\newblock \emph{Found. Trends Theor. Comput. Sci.}, 9\penalty0 (3-4):\penalty0
  211--407, 2014.

\bibitem[Enthoven \& Al{-}Ars(2020)Enthoven and Al{-}Ars]{enthoven2020}
Enthoven, D. and Al{-}Ars, Z.
\newblock An overview of federated deep learning privacy attacks and defensive
  strategies.
\newblock \emph{arXiv preprint arXiv:2004.04676}, 2020.

\bibitem[Go et~al.(2009)Go, Bhayani, and Huang]{sent140}
Go, A., Bhayani, R., and Huang, L.
\newblock Twitter sentiment classification using distant supervision.
\newblock \emph{Stanford Tech. Report}, 2009.

\bibitem[Gopi et~al.(2021)Gopi, Lee, and Wutschitz]{gopi2021numerical}
Gopi, S., Lee, Y.~T., and Wutschitz, L.
\newblock Numerical composition of differential privacy.
\newblock \emph{arXiv preprint arXiv:2106.02848}, 2021.

\bibitem[Gu et~al.(2020)Gu, Mao, Han, Liu, Wu, Yu, Finnie, Yu, Zhai, and
  Zukoski]{Gu+20}
Gu, X., Mao, Y., Han, J., Liu, J., Wu, Y., Yu, C., Finnie, D., Yu, H., Zhai,
  J., and Zukoski, N.
\newblock Generating representative headlines for news stories.
\newblock In \emph{Proc. of Intern. Conf. WWW'20}, 2020.

\bibitem[He et~al.(2020)He, Li, So, Zeng, Zhang, Wang, Wang, Vepakomma, Singh,
  Qiu, Zhu, Wang, Shen, Zhao, Kang, Liu, Raskar, Yang, Annavaram, and
  Avestimehr]{He+20}
He, C., Li, S., So, J., Zeng, X., Zhang, M., Wang, H., Wang, X., Vepakomma, P.,
  Singh, A., Qiu, H., Zhu, X., Wang, J., Shen, L., Zhao, P., Kang, Y., Liu, Y.,
  Raskar, R., Yang, Q., Annavaram, M., and Avestimehr, S.
\newblock Fedml: A research library and benchmark for federated machine
  learning.
\newblock \emph{arXiv preprint arXiv:2007.13518}, 2020.

\bibitem[He et~al.(2021)He, Balasubramanian, Ceyani, Yang, Xie, Sun, He, Yang,
  Yu, Rong, Zhao, Huang, Annavaram, and Avestimehr]{He+21}
He, C., Balasubramanian, K., Ceyani, E., Yang, C., Xie, H., Sun, L., He, L.,
  Yang, L., Yu, P.~S., Rong, Y., Zhao, P., Huang, J., Annavaram, M., and
  Avestimehr, S.
\newblock Fedgraphnn: A federated learning system and benchmark for graph
  neural networks.
\newblock \emph{arXiv preprint arXiv:2104.07145}, 2021.

\bibitem[Houlsby et~al.(2019)Houlsby, Giurgiu, Jastrzebski, Morrone,
  De~Laroussilhe, Gesmundo, Attariyan, and Gelly]{Houlsby+19}
Houlsby, N., Giurgiu, A., Jastrzebski, S., Morrone, B., De~Laroussilhe, Q.,
  Gesmundo, A., Attariyan, M., and Gelly, S.
\newblock Parameter-efficient transfer learning for {NLP}.
\newblock In \emph{Proc. of the 36th Intern. Conf. on Machine Learning}, pp.\
  2790--2799, 2019.

\bibitem[Jhunjhunwala et~al.(2021)Jhunjhunwala, Gadhikar, Joshi, and
  Eldar]{JGJE21}
Jhunjhunwala, D., Gadhikar, A., Joshi, G., and Eldar, Y.~C.
\newblock Adaptive quantization of model updates for communication-efficient
  federated learning.
\newblock \emph{arXiv preprpint arXiv:2102.04487}, 2021.

\bibitem[Karimireddy et~al.(2020)Karimireddy, Kale, Mohri, Reddi, Stich, and
  Suresh]{Karimireddy+20}
Karimireddy, S.~P., Kale, S., Mohri, M., Reddi, S., Stich, S., and Suresh,
  A.~T.
\newblock {SCAFFOLD}: Stochastic controlled averaging for federated learning.
\newblock In \emph{Proc. of the 37th Inter. Conf. on Machine Learning}, pp.\
  5132--5143, 2020.

\bibitem[Kingma \& Ba(2017)Kingma and Ba]{KiBa17}
Kingma, D.~P. and Ba, J.
\newblock Adam: A method for stochastic optimization.
\newblock \emph{arXiv preprint arXiv:a1412.6980}, 2017.

\bibitem[Konecny et~al.(2015)Konecny, McMahan, and Ramage]{KMR15}
Konecny, J., McMahan, B.~H., and Ramage, D.
\newblock {Federated Optimization: Distributed Optimization Beyond the
  Datacenter}.
\newblock \emph{arXiv preprint arXiv:1511.03575v1}, 2015.

\bibitem[LeCun \& Cortes(2010)LeCun and Cortes]{mnist2010}
LeCun, Y. and Cortes, C.
\newblock {MNIST} handwritten digit database.
\newblock 2010.
\newblock URL \url{http://yann.lecun.com/exdb/mnist/}.

\bibitem[Li et~al.(2019{\natexlab{a}})Li, Wen, and He]{LWH19}
Li, Q., Wen, Z., and He, B.
\newblock Practical federated gradient boosting decision trees.
\newblock \emph{arXiv preprint arXiv:1911.04206}, 2019{\natexlab{a}}.

\bibitem[Li et~al.(2019{\natexlab{b}})Li, Sahu, Talwalkar, and Smith]{LSTS18}
Li, T., Sahu, A.~K., Talwalkar, A., and Smith, V.
\newblock {Federated Learning: Challenges, Methods, and Future Directions}.
\newblock \emph{arXiv preprint arXiv:1908.07873v1}, 2019{\natexlab{b}}.

\bibitem[Li et~al.(2019{\natexlab{c}})Li, Huang, Yang, Wang, and Zhang]{Li+19}
Li, X., Huang, K., Yang, W., Wang, S., and Zhang, Z.
\newblock On the convergence of fedavg on non-iid data.
\newblock \emph{arXiv preprint arXiv:1907.02189}, 2019{\natexlab{c}}.

\bibitem[Lian et~al.(2015)Lian, Huang, Li, and Liu]{LHLL15}
Lian, X., Huang, Y., Li, Y., and Liu, J.
\newblock Asynchronous parallel stochastic gradient for nonconvex optimization.
\newblock In \emph{Proc. of 28th Intl. Conf. on Advances in Neural Information
  Processing Systems (NIPS 2015)}, 2015.

\bibitem[Liang et~al.(2020)Liang, Javid, Skoglund, and Chatterjee]{LJSC20}
Liang, X., Javid, A.~M., Skoglund, M., and Chatterjee, S.
\newblock {Asynchronous Decentralized Learning of a Neural Network}.
\newblock \emph{arXiv preprint arXiv:2004.05082v1}, 2020.

\bibitem[Liu et~al.(2022{\natexlab{a}})Liu, Xu, and Wang]{LXW22}
Liu, P., Xu, X., and Wang, W.
\newblock Threats, attacks and defenses to federated learning: issues, taxonomy
  and perspectives.
\newblock \emph{Cybersecurity}, 5, Feb. 2022{\natexlab{a}}.

\bibitem[Liu et~al.(2022{\natexlab{b}})Liu, Shi, Xie, Li, Hu, Kim, Xu, Li, and
  Song]{https://doi.org/10.48550/arxiv.2207.10308}
Liu, X., Shi, T., Xie, C., Li, Q., Hu, K., Kim, H., Xu, X., Li, B., and Song,
  D.
\newblock Unifed: A benchmark for federated learning frameworks,
  2022{\natexlab{b}}.
\newblock URL \url{https://arxiv.org/abs/2207.10308}.

\bibitem[McMahan et~al.(2017)McMahan, Moore, Ramage, Hampson, and
  Arcas]{McMahan+17}
McMahan, H.~B., Moore, E., Ramage, D., Hampson, S., and Arcas, B.
\newblock {Communication-efficient Learning of Deep Networks from Decentralized
  Data}.
\newblock In \emph{Proc. of Intern. Conf. on Artificial Intelligence and
  Statistics}, pp.\  1273--–1282, 2017.

\bibitem[Mironov(2017)]{mironov2017renyi}
Mironov, I.
\newblock R{\'e}nyi differential privacy.
\newblock In \emph{2017 IEEE 30th Computer Security Foundations Symposium
  (CSF)}, pp.\  263--275. IEEE, 2017.

\bibitem[Nvidia()]{nvidia}
Nvidia.
\newblock Nvidia/nccl: Optimized primitives for collective multi-gpu
  communication.
\newblock URL \url{https://github.com/NVIDIA/nccl}.

\bibitem[Panayotov et~al.(2015)Panayotov, Chen, Povey, and Khudanpur]{PCPK15}
Panayotov, V., Chen, G., Povey, D., and Khudanpur, S.
\newblock {LibriSpeech: an ASR corpus based on public domain audio books}.
\newblock In \emph{Proc. of Intern. Conf. on Acoustics, Speech and Signal
  Processing}, 2015.

\bibitem[Patarasuk \& Yuan(2009)Patarasuk and Yuan]{PaY09}
Patarasuk, P. and Yuan, X.
\newblock {Bandwidth Optimal All-reduce Algorithms for Clusters of
  Workstations}.
\newblock \emph{J. Parallel Distrib. Comput.}, 69\penalty0 (2):\penalty0
  117--124, 2009.

\bibitem[Reddi et~al.(2020)Reddi, Charles, Zaheer, Garrett, Rush,
  Kone{\v{c}}n{\`y}, Kumar, and McMahan]{reddi2020adaptive}
Reddi, S., Charles, Z., Zaheer, M., Garrett, Z., Rush, K., Kone{\v{c}}n{\`y},
  J., Kumar, S., and McMahan, H.~B.
\newblock Adaptive federated optimization.
\newblock \emph{arXiv preprint arXiv:2003.00295}, 2020.

\bibitem[Reddi et~al.(2021)Reddi, Charles, Zaheer, Garrett, Rush,
  Kone{\v{c}}n{\'y}, Kumar, and McMahan]{Reddi+21}
Reddi, S.~J., Charles, Z., Zaheer, M., Garrett, Z., Rush, K.,
  Kone{\v{c}}n{\'y}, J., Kumar, S., and McMahan, H.~B.
\newblock Adaptive federated optimization.
\newblock In \emph{9th Intern. Conf. on Learning Representations, {ICLR-21}},
  May 2021.

\bibitem[Rosenblatt(1958)]{Rosenblatt58}
Rosenblatt, F.
\newblock The perceptron: A probabilistic model for information storage and
  organization in the brain.
\newblock \emph{Psychological review}, 65, 1958.

\bibitem[Seide et~al.(2014)Seide, Fu, Droppo, Li, and Yu]{seide20141}
Seide, F., Fu, H., Droppo, J., Li, G., and Yu, D.
\newblock 1-bit stochastic gradient descent and its application to
  data-parallel distributed training of speech dnns.
\newblock In \emph{15th Annual Conf. of the Intern. Speech Communication
  Association}, 2014.

\bibitem[Sergeev \& Bals(2018)Sergeev and Bals]{SeBa18}
Sergeev, A. and Bals, M.~D.
\newblock {Horovod: Fast and Easy Distributed Deep Learning in TensorFlow}.
\newblock \emph{arXiv preprint arXiv:1802.05799}, 2018.

\bibitem[Shamir et~al.(2013)Shamir, Srebro, and Zhang]{SSZ13}
Shamir, O., Srebro, N., and Zhang, T.
\newblock {Communication Efficient Distributed Optimization Using an
  Approximate Newton-type Method}.
\newblock \emph{arXiv preprint arXiv:1312.7853}, 2013.

\bibitem[Strom(2015)]{strom2015scalable}
Strom, N.
\newblock Scalable distributed dnn training using commodity gpu cloud
  computing.
\newblock In \emph{16th Annual Conf. of the Intern. Speech Communication
  Association}, 2015.

\bibitem[Sun et~al.(2022)Sun, Salim, and
  Richtárik]{https://doi.org/10.48550/arxiv.2206.00920}
Sun, L., Salim, A., and Richtárik, P.
\newblock Federated learning with a sampling algorithm under isoperimetry,
  2022.
\newblock URL \url{https://arxiv.org/abs/2206.00920}.

\bibitem[Tang et~al.(2015)Tang, Qin, and Liu]{yelp2015}
Tang, D., Qin, B., and Liu, T.
\newblock Document modeling with gated recurrent neural network for sentiment
  classification.
\newblock In \emph{Proc. of ACL Intern. Conf. EMNLP'15}, 2015.

\bibitem[Wang et~al.(2022)Wang, Dimitriadis, Koyejo, and
  Tople]{https://doi.org/10.48550/arxiv.2210.01834}
Wang, X., Dimitriadis, D., Koyejo, S., and Tople, S.
\newblock Invariant aggregator for defending federated backdoor attacks, 2022.
\newblock URL \url{https://arxiv.org/abs/2210.01834}.

\bibitem[Wangni et~al.(2018)Wangni, Wang, Liu, and Zhang]{wangni2018gradient}
Wangni, J., Wang, J., Liu, J., and Zhang, T.
\newblock Gradient sparsification for communication-efficient distributed
  optimization.
\newblock \emph{Advances in Neural Information Processing Systems},
  31:\penalty0 1299--1309, 2018.

\bibitem[Wei et~al.(2020)Wei, Li, Ding, Ma, Yang, Farokhi, Jin, Quek, and
  Poor]{wei2020}
Wei, K., Li, J., Ding, M., Ma, C., Yang, H.~H., Farokhi, F., Jin, S., Quek, T.
  Q.~S., and Poor, H.~V.
\newblock Federated learning with differential privacy: Algorithms and
  performance analysis.
\newblock \emph{IEEE Transactions on Information Forensics and Security},
  15:\penalty0 3454--3469, 2020.
\newblock \doi{10.1109/TIFS.2020.2988575}.

\bibitem[Wolford(2021)]{GDPR}
Wolford, B.
\newblock {A Guide to GDPR Data Privacy Requirements}.
\newblock \url{https://gdpr.eu/data-privacy/}, 2021.

\bibitem[You et~al.(2017)You, Gitman, and Ginsburg]{YGG17}
You, Y., Gitman, I., and Ginsburg, B.
\newblock Large batch training of convolutional networks.
\newblock \emph{arXiv preprint arXiv:1708.03888}, 2017.

\bibitem[You et~al.(2020)You, Li, Reddi, Hseu, Kumar, Bhojanapalli, Song,
  Demmel, Keutzer, and Hsieh]{You+20}
You, Y., Li, J., Reddi, S., Hseu, J., Kumar, S., Bhojanapalli, S., Song, X.,
  Demmel, J., Keutzer, K., and Hsieh, C.-J.
\newblock {Large Batch Optimization for Deep Learning: Training BERT in 76
  minutes}.
\newblock \emph{arXiv preprint arXiv:1904.00962}, 2020.

\bibitem[Zhang et~al.(2021)Zhang, Chen, Cheng, Binh, and Yu]{zhang2021}
Zhang, J., Chen, B., Cheng, X., Binh, H. T.~T., and Yu, S.
\newblock {PoisonGAN: Generative Poisoning Attacks Against Federated Learning
  in Edge Computing Systems}.
\newblock \emph{IEEE Internet of Things Journal}, 8\penalty0 (5):\penalty0
  3310--3322, 2021.

\bibitem[Zhao et~al.(2018)Zhao, Li, Lai, Suda, Civin, and
  Chandra]{zhao2018federated}
Zhao, Y., Li, M., Lai, L., Suda, N., Civin, D., and Chandra, V.
\newblock Federated learning with non-iid data.
\newblock \emph{arXiv preprint arXiv:1806.00582}, 2018.

\bibitem[Zhu et~al.(2021)Zhu, Dong, and
  Wang]{DBLP:journals/corr/abs-2106-08567}
Zhu, Y., Dong, J., and Wang, Y.
\newblock Optimal accounting of differential privacy via characteristic
  function.
\newblock \emph{arXiv preprint arXiv:2106.08567}, 2021.

\end{thebibliography}


\begin{thebibliography}{8}
\providecommand{\natexlab}[1]{#1}
\providecommand{\url}[1]{\texttt{#1}}
\expandafter\ifx\csname urlstyle\endcsname\relax
  \providecommand{\doi}[1]{doi: #1}\else
  \providecommand{\doi}{doi: \begingroup \urlstyle{rm}\Url}\fi

\bibitem[Author(2018)]{anonymous}
Author, N.~N.
\newblock Suppressed for anonymity, 2018.

\bibitem[Duda et~al.(2000)Duda, Hart, and Stork]{DudaHart2nd}
Duda, R.~O., Hart, P.~E., and Stork, D.~G.
\newblock \emph{Pattern Classification}.
\newblock John Wiley and Sons, 2nd edition, 2000.

\bibitem[Kearns(1989)]{kearns89}
Kearns, M.~J.
\newblock \emph{Computational Complexity of Machine Learning}.
\newblock PhD thesis, Department of Computer Science, Harvard University, 1989.

\bibitem[Langley(2000)]{langley00}
Langley, P.
\newblock Crafting papers on machine learning.
\newblock In Langley, P. (ed.), \emph{Proceedings of the 17th International
  Conference on Machine Learning (ICML 2000)}, pp.\  1207--1216, Stanford, CA,
  2000. Morgan Kaufmann.

\bibitem[Michalski et~al.(1983)Michalski, Carbonell, and
  Mitchell]{MachineLearningI}
Michalski, R.~S., Carbonell, J.~G., and Mitchell, T.~M. (eds.).
\newblock \emph{Machine Learning: An Artificial Intelligence Approach, Vol. I}.
\newblock Tioga, Palo Alto, CA, 1983.

\bibitem[Mitchell(1980)]{mitchell80}
Mitchell, T.~M.
\newblock The need for biases in learning generalizations.
\newblock Technical report, Computer Science Department, Rutgers University,
  New Brunswick, MA, 1980.

\bibitem[Newell \& Rosenbloom(1981)Newell and Rosenbloom]{Newell81}
Newell, A. and Rosenbloom, P.~S.
\newblock Mechanisms of skill acquisition and the law of practice.
\newblock In Anderson, J.~R. (ed.), \emph{Cognitive Skills and Their
  Acquisition}, chapter~1, pp.\  1--51. Lawrence Erlbaum Associates, Inc.,
  Hillsdale, NJ, 1981.

\bibitem[Samuel(1959)]{Samuel59}
Samuel, A.~L.
\newblock Some studies in machine learning using the game of checkers.
\newblock \emph{IBM Journal of Research and Development}, 3\penalty0
  (3):\penalty0 211--229, 1959.

\end{thebibliography}


\newpage
